\documentclass[10pt,twocolumn,twoside] {IEEEtran}
\usepackage{graphicx,amsfonts}
\usepackage{subfig,color}
\usepackage{cite}
\ifCLASSINFOpdf
\else
\fi
\usepackage[cmex10]{amsmath}
\usepackage{array,multirow}
\usepackage{mdwmath}
\usepackage{mdwtab}
\begin{document}

\title{Road Friction Estimation for Connected Vehicles using Supervised Machine Learning }

\author{Ghazaleh Panahandeh$^\diamond$, Erik Ek$^*$, Nasser Mohammadiha$^{\diamond*}$\\$^\diamond$ Zenuity AB, $^*$Chalmers University of Technology
\thanks{This paper is published at IEEE Intelligent Vehicles Symposium (IV), 2017: http://ieeexplore.ieee.org/document/7995885/}
}

\markboth{Published at IEEE Intelligent Vehicles Symposium (IV), 2017}%
{Published at IEEE Intelligent Vehicles Symposium (IV), 2017}

\maketitle
\begin{abstract}
\thispagestyle{empty}
In this paper, the problem of road friction prediction from a fleet of connected vehicles is investigated. A framework is proposed to predict the road friction level using both historical friction data from the connected cars and data from weather stations, and comparative results from different methods are presented.
The problem is formulated as a classification task where the available data is used to train three machine learning models including logistic regression, support vector machine, and neural networks to predict the friction class (slippery or non-slippery) in the future for specific road segments.
In addition to the friction values, which are measured by moving vehicles, additional parameters such as humidity, temperature, and rainfall are used to obtain a set of descriptive feature vectors as input to the classification methods.
The proposed prediction models are evaluated for different prediction horizons (0 to 120 minutes in the future) where the evaluation shows that the neural networks method leads to more stable results in different conditions.

\end{abstract}
\section{Introduction}
Connected vehicle technology is foreseen to play an important role in reducing the number of traffic accidents while being one of the main enabling components for autonomous driving. One of the application of such connection is to provide accurate information about the road condition such as friction level to drivers or the intelligent systems controlling the car.
Road surface friction can be defined as the grip between car tyre and underlying surface. 
During winter times when the temperature decreases dramatically, friction level reduces substantially, which can increase the risk of car accidents.
Studies indicate that road conditions such as surface temperature, type of road, and structure of the road sides play an important role in the measured friction level, and some of these conditions can vary significantly within short distances under specific weather situations.
Road friction prediction based on the past sensor measurements available in
the cars, e.g., temperature and sun light, has advantages of being independent of the road structure and surrounding infrastructure. 

Intelligent forecast systems rely on the availability of high quality data in order to allow their multiple actors to make correct decisions in diverse traffic situations. These systems have the potential to increase the safety of roads users by means of the timely sharing of road-related information. 
With the advances in car-to-car communication technology, today, Volvo cars are equipped with slippery road condition warning system to improve road safety and traffic flow. In the current system, real-time data are transmitted to the cloud and once slippery road condition is detected this information will be distributed to nearby cars.

Solutions to the problem of road condition estimation can be divided into two groups. The first group mainly focuses on on-board road condition estimation without having any forecasting model.
For instance tire-road friction estimation that can be achieved by measuring vehicle dynamic response~\cite{gustafsson1997slip,rajamani2012algorithms,shao2016road}.
The second group, on the other hand, focuses on building predictive models to estimate road conditions ahead of time using historical data~\cite{berrocal2012probabilistic,venkadesh2013genetic, marchetti2015methodology,nuijten2016runway}.
For example in~\cite{berrocal2012probabilistic} the slippery road condition is estimated using Markov chain Monte Carlo method. The probability of ice formation is determined by constructing the joint probability distribution of temperature and precipitation which results in probability of ice formation. 
Similarly a temperature predictive model is introduced in~\cite{venkadesh2013genetic} using artificial neural network (ANN) by performing a generic algorithm search to determine the optimal duration and resolution of prior data for each weather variable that was considered as a potential input variable for the model.
These predictive models are mainly based on using data from weather stations, where the drawback is that weather stations can not provide the total map of road network conditions as they are located mainly on the main roads and usually in far distance from each other. Hence, providing the full map of entire road condition is highly challenging~\cite{mahoney2013realizing,dey2015potential,karsisto2016using}.

In this paper, we propose a third group where both data from weather stations and the historical friction data from fleets of connected cars are used to predict the slippery road condition in the future. The proposed approach can easily be integrated into the available connected vehicle technology to improve traffic flow management by notifying road administration by the need to handle slippery road segments. Another potential service is slippery road warning to vehicles/drivers, where slippery road segments are identified. 

More specifically, in this paper, statistical relationship between the measured friction levels and the available explanatory variables are investigated and proper pre-processing and feature construction methods are proposed in order to train supervised machine learning methods to predict the friction level at a specific road segment for the time horizon of 0 to 120 minutes in the future. The problem is formulated as a binary classification (slippery or non-slippery) task and three classification methods are implemented to solve the task. The main contribution of this work can be seen as the proposed framework and performing a comparative study of the available methods to solve the prediction task. Our experiments show that low error rates in the order of 20-30 $\%$ can be obtained while slightly different results are obtained for different road segments and different prediction horizons.

The structure of this paper is as follows. 
Data acquisition and data sets are introduced in Section~\ref{S:dataAq}. The problem description and the proposed methods are presented in Section~\ref{S:method}. 
Experimental validations and discussions are reported
in Section~\ref{S:result}. Finally, the conclusion of the study is given in Section~\ref{S:conclusion}.

\section{Data acquisition}\label{S:dataAq}
In this section, we will first introduce the structure of our data set. Then in the second part, the pre-processing phase is described.

\subsection{Data Set}
\begin{figure}[t]
\centering
\includegraphics[width=.9\linewidth]{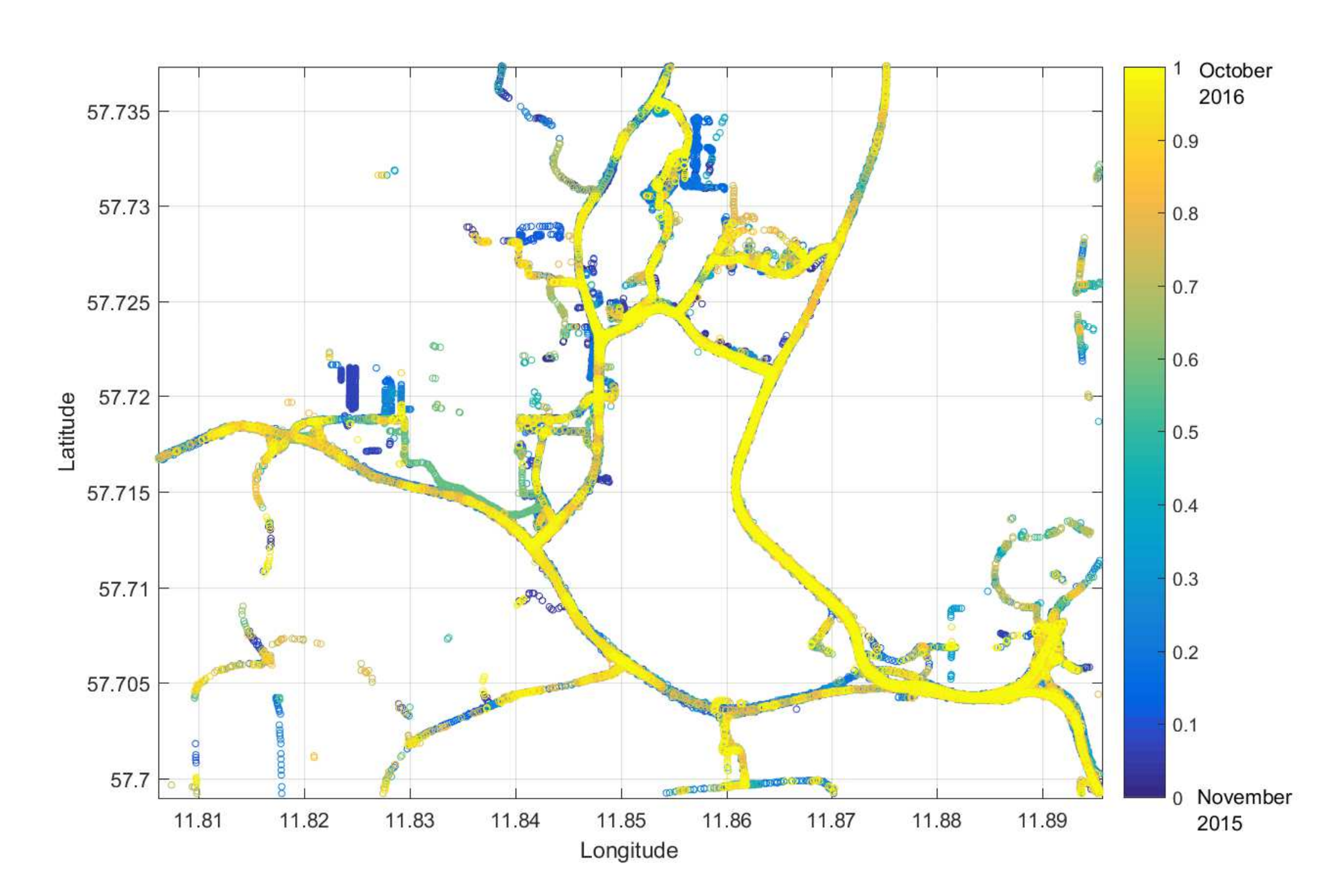}
\caption{The graph of friction value measurements during the time period November 2015 to October 2016. The measurements are color coded with respect to time where blue dots represent measurements taken in the beginning and and the yellow dots are measurements taken more recently. Some of the earlier indicators are not seen since they are hidden behind more recent measured data.}\label{fig:regions}
\end{figure}

The road friction prediction is achieved by using the measurements from fleet of connected Volvo cars in the city of Gothenburg in Sweden between the period of November 2015 until October 2016. Fig.~\ref{fig:regions} shows the geo-location of the measured friction values which are used in our prediction models. In this figure, data is color coded with respect to time for the considered period of time.

\begin{figure}[t]
\centering
\includegraphics[width=.9\linewidth]{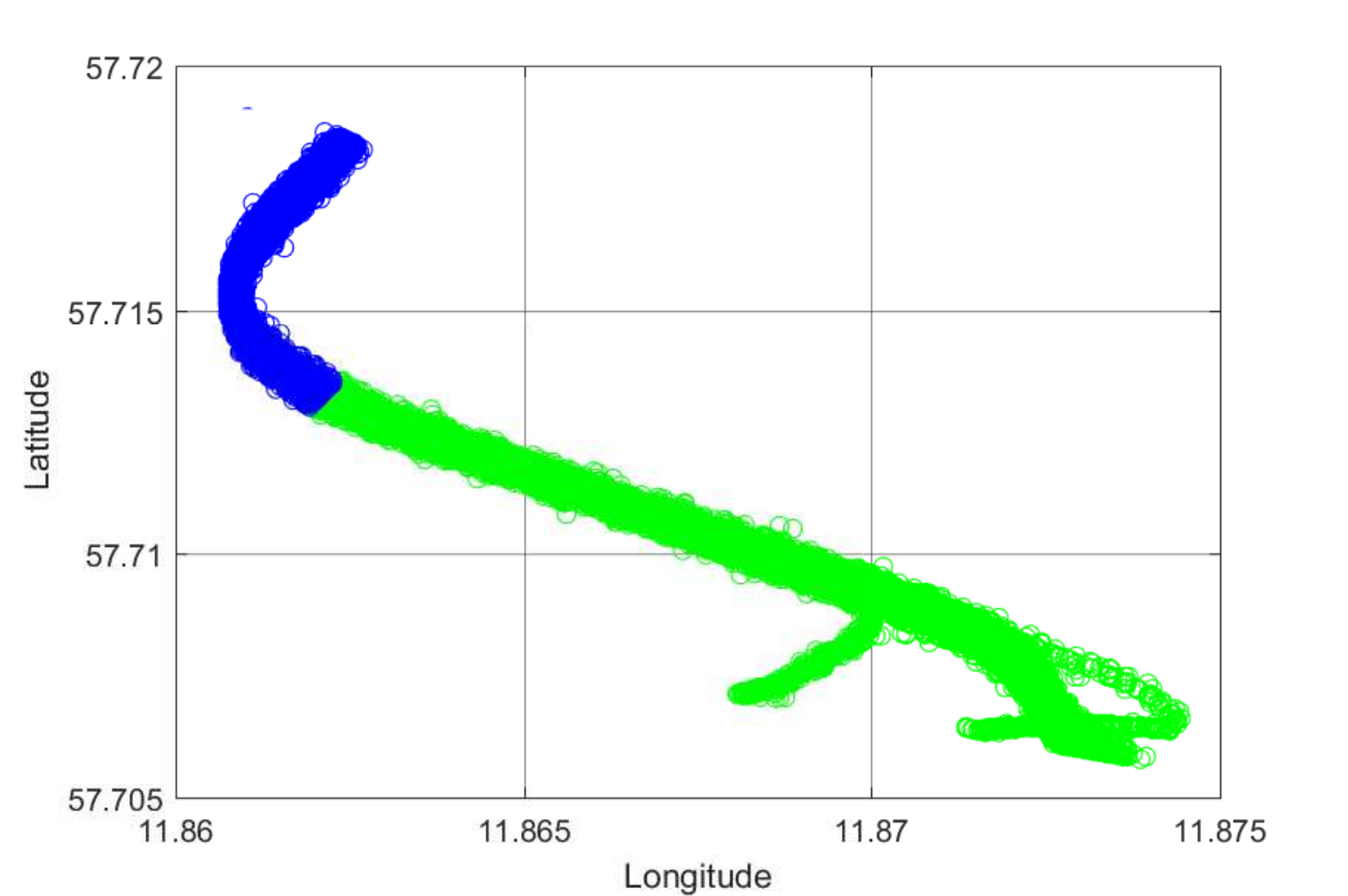}
\caption{The position of 9281 friction measurements during the period of November 2015 to October 2016 is plotted for two road segments.}\label{fig:roadsegment}
\end{figure}

From fleet of connected cars, the following measurements are used in our prediction model: estimated friction values and their corresponding confidence/quality of measurement, wiper speed, ambient temperature, time stamp, and the road segment id where the measurement accrued.
The friction values are calculated by estimating the wheel slip computed from sensor signals and each value is labeled with a predefined threshold value indicating if the road is slippery or not. This confidence level is defined on an ordinal scale. It is worth mentioning that those friction values whose confidence level are below a given threshold are not used for the analysis.
In addition to the measured data from the vehicles, weather data collected from Swedish meteorological and hydrological institute (SMHI) stations and Swedish road administration are also used in this paper. These stations are located in three different locations in the city of Gothenburg (Lindholmen, S\"ave and Gothenburg city center). The weather data consists of data points measured at 30-minute intervals and the following fields are used from the stations: air temperature, road surface temperature, humidity, dewpoint temperature, rainfall, snowfall, and windspeed. The amount of rain and snowfall is measured in millimeter per 30 minutes and is collected six meters above the ground. The temperatures and humidity are measured two meter above the ground. In this dataset the humidity variable takes values between zero and one.
To better represent density of data in road segments, an example of collected measurement in two connected road segments is plotted in Fig.~\ref{fig:roadsegment}.


\subsection{Pre-processing}
We quantize the time into discrete intervals, i.e., the continuous time is quantized to fall into a finite set of intervals. By averaging the friction values from all measurements within these intervals, we get a down-sampled representation of the events. The main reason to do this is that sometimes several measurements are taken within a very short period of time which are highly correlated. In the evaluation section, we investigate the effect of the interval length in the prediction performance.

Due to the stochastic nature of the measurements, a suitable pre-processing method has to be used before extracting features in order to, e.g., remove outliers and other non-consistent data points. 
Moreover by plotting histograms, we analysed the data distribution, e.g., if there are a few distinct values or if the the data distribution is Gaussian or heavy tailed distributions.
We aim to develop a friction prediction for both the current time as well as in the future, where the history data and previously measured friction values are also used to build a model.
In our setup, measurement data within four hours are included in the feature set. These past friction values are stored in the data set along with the distance and duration from the initial measurement, our responsive variable, and for each data sample, the previously measured friction values are weighted with linearly decreasing weights with respect to the distance and duration from the initial friction measurement in order to extract feature vectors. It is worth mentioning that measurements taken after the responsive variable can not be used in the prediction.
For each road segment, we include measurements in the radios of 3.334 km in the data set for that segment, and thus samples in  data are not strictly limited to the specific road segments in order to get variety in the collected data. The choice of this radius was based on field knowledge in order to incorporate a major part of the regions we are studying.
The final step before training, the models is normalization of the data. The created data set will be normalized so that features corresponding to each sample have zero mean and unit variance. By adjusting values measured on different scales to a common scale each feature will be of same importance, which is particularly important for PCA.

As number of passing cars in a street or road might significantly vary depending on the weather condition, road connections etc, the number of received measurements also vary. A related issue arises when the distribution of the passing cars during the
desired interval, e.g., one hour, is different over time. These issues lead to heterogeneous structure in
the input data, and this has to be fixed before processing. Herein, we solved the problem in a case by case basis, we down-sampled the data when there are many measurements. However, we did not consider the problem of missing data, where there is too few measurements within a desired time interval. In this case, imputation methods~\cite{Schafer2202} can be utilized and as a feature work we are currently looking into this problem.


\section{Method Description}\label{S:method}
Our goal with this study is to compare a set of state-of-the-art classification methods in order to predict the slippery road condition. The overall procedure after collecting the dataset and pre-processing can be divided in the following steps: 1) dimensionality reduction, 2) training the supervised classifiers, 3) evaluation of the learned models. The first two steps are discussed in this section and then the final step is covered in Section~\ref{S:result}.

\subsection{Dimensionality Reduction}
Since our data sets contain an abundant amount of features compared to the number of samples, we need to reduce the dimensionality of the data set. Hence, the space complexity of the data set is reduced which can lead the structure of the data to be more interpretable to the classification models. In this paper, we use the PCA method for dimentionality reduction. This method is based upon the variation of the data itself, which finds orthogonal basis of the data with basis vectors that are sorted by the amount of variance in their direction. With this technique we can extract the subspaces that accounts for a known amount of variance. We can set a threshold for the amount of variance that we want to extract or just set a maximum number of dimensions. PCA extracts a linear subspace which we then use as as the feature set.
Our assumption is that subspaces with lower variances will represent the information which are not changing as much as the other subspaces and might therefore be less important. Finding a good threshold between useful data and noise varies from case to case. In this analysis we'll use a combination of looking at the results and using t-SNE to find a suitable maximum number of dimensions or a threshold for the variance.
In this project we will apply the PCA algorithm before we train the logistic regression (LR) and the support vector machine (SVM) models. In principle, the linear transformation performed by PCA can be performed just as well by the input layer weights of the neural network, so it isn't strictly speaking necessary to use PCA for the neural network. However, as the number of weights in the network increases, the amount of data needed to be able to reliably determine the weights of the network also increases and overfitting becomes more of an issue.\\
The benefit of dimensionality reduction is that it reduces the size of the network, and hence the amount of data needed to train it. The disadvantage of using PCA is that the discriminative information that distinguishes one class from another might be in the low variance components, so using PCA can make performance worse when training the neural network.

\subsection{Overview of Training and Classification Methods}
\begin{figure}
\centering
\includegraphics[width=.8\columnwidth]{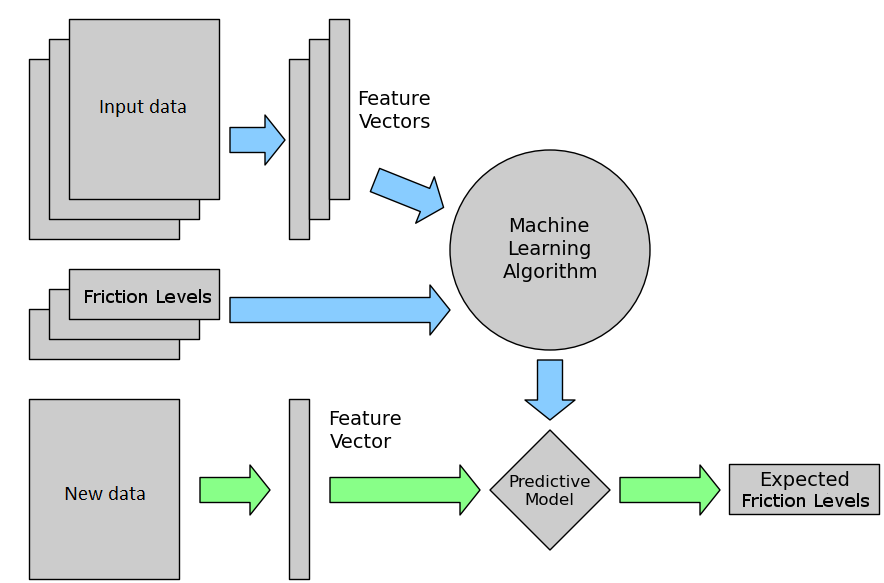}
\caption{A supervised learning algorithm analyzes the training data and produces an predictive model, which can be used to predict the friction level. An optimal scenario will allow for the algorithm to correctly determine the class labels for unseen instances.}\label{fig:supervised_class}
\end{figure}

Herein three different classification methods are evaluated for the purpose of slippery road prediction, namely LR, SVM, and multi-layer neural networks.
We formulate the prediction problem as a supervised binary classification where the aim is to predict the friction level if is low or high in a specific road segment. 
Supervised machine learning is the search for algorithms that reason from externally supplied instances
to produce general hypotheses, which then make predictions about future instances~\cite{bishop2006}. In other words, the
goal of supervised learning is to build a concise model of the distribution of class labels in terms of
predictor features. The resulting classifier is then used to assign class labels to the testing instances
where the values of the predictor features are known, but the value of the class label is unknown. In our case, the goal is to predict the class labels "high friction" or "low friction". Fig.\ref{fig:supervised_class} shows the overall process of using supervised classification in this context. 

\thispagestyle{empty}
By varying the complexity of the classification algorithms we get a better understanding of the amount of complexity in the data sets. We use the LR because it is an intrinsically simple algorithm which has a low variance and so is less prone to overfitting. Then we have the SVM which is very different from LR. The advantage of SVM is that it has techniques for improving the regularisation. Also, using kernel SVM version, we can model nonlinear relations. The third algorithm is the ANN, which usually performs well when it is a priori difficult or impractical to formulate specific assumptions about a possible nonlinear relationship of several variables or when a precise knowledge of the relationship requires some functional approximation.

For each available labeling, which is denoted by $y_{t}$ at time $t$ and
shows the friction level of the road, we structure the received
measurements from $t-T$ to $t$ to build a $df_{s}$-dimensional
vectors $\mathbf{z}_{t}$. These two make an input-output pair and it has to be noted that given this pair, the rest of the algorithm
does not depend on how this two are coupled. In the following, a short overview of the considered classification methods is given.

\subsubsection{Logistic Regression}
To begin with, we implemented LR through Matlab's built-in function glmfit. LR is a technique in statistics to train models that aimes at find the best fitting, most accurate and sensible model to assess the relationship between a set of responsive variables and at least one explanatory variable. It differs from the linear regression in that it can be applied when the dependent variable is categorical and does not require rigorous assumptions to be met. LR is a prognostic model that is fitted where there is a binary dependent variable. The categories we are trying to predict are low and high friction, which are coded as 0 and 1. It results is a straightforward interpretation. We use the logit transformation which is referred to as the link function in LR. Although the dependent variable in LR is binomial, the logit is the continuous criterion upon which linear regression is conducted. 

\subsubsection{Support Vector Machine}
The SVM method differs from the LR in that it is a discriminative classifier formally defined by a separating hyperplane. From labeled training data, the algorithm outputs an optimal hyperplane which categorizes new examples. The reason we consider SVM is because it can be used as a non-probabilistic binary nonlinear classifier and is a very popular technique for supervised classification. One of the big advantages of SVM is that it is still effective in cases where number of dimensions is greater than the number of samples. It also performs good in high dimensional spaces and the algorithm is very versatile.

SVM is a discriminative classifier formally defined by a separating hyperplane. From labeled training data, the algorithm outputs an optimal hyperplane which categorizes new examples. Here, soft-margin is used to improve the results while we also use radial basis function to build a Kernel SVM.

Here, we train SVM with only two classes. When using the SVM we can create a nonlinear decision boundary by transforming the data by a non-linear function to a higher dimensional space. The data points are then moved from the original space $I$ onto a feature space we call $F$. 
The goal of the algorithm is to find a hyperplane which is represented with the equation $w^Tx+b=0$ where $w\in \mathbb{R}^d$. The hyperplane determines the margin between the classes. Ideally we are looking for a hyperplane that can separate the two classes of data by maximizing the distance from the closest support vectors from each class of the hyperplane is equal; thus the constructed hyperplane searches for the maximal margin. One trick consist of using a soft margin to prevent overfitting noisy data. It's a regularization technique that controls the trade-off between maximizing the margin and minimizing the training error.
In this case a slack variable $\xi_i$ is introduced that allows some data points to lie within the margin, and a constant $C>0$ which determines the trade-off between maximizing the margin and the number of support vectors within that margin.
The following loss function is minimized to train SVM:
\begin{equation}
  \frac{||w||^2}{2}+C\sum^n_{i=1}\xi_i  
\end{equation}
where constraints $y_i(w^T\phi(x_i)+b)\geq1-\xi_i$ $\forall i=1,\dots,n$ and $\xi_i\geq0$ $\forall i=1,\dots,n$ are applied.
The minimization problem is solved using Lagrange multipliers which results in
\begin{equation}
f(x)=\text{sign}\left(\sum_{i=1}^n \alpha_i y_i K(x,x_i)+b\right)
\end{equation}
where $K(\cdot,\cdot)$ represents the kernel, \text{sign} is the sign function, and $\alpha_i$ are the Lagrange multipliers. SVM is spares and therefore there will be only few Lagrange multipliers with non-zero values. We skip the details of the kernel SVM and instead refer to the references, e.g., \cite{bishop2006}.

\subsubsection{Artificial Neural Networks}
Neural networks with many hidden layers have been very successful in recent years and different architectures are proposed to solve a various range of problems. In our study, we investigated the use of feed forward neural networks and using the experiments studied the effect of the number of hidden layers in the performance.

Before each training session, all weights were set to a random number between -0.1 and 0.1. We found that Relu activation function~\cite{glorot2011deep} works best with our data sets. This activation function can be motivated by enabling an efficient gradient propagation. Meaning that we get no vanishing or exploding gradient problems. The function is scale-invariant and enables sparse activation, which means that in a randomly initialized network, only about half of the hidden units are activated.
This is possible since the activation function is not continuous below zero which blocks the gradient. This is in fact part of the rectifiers advantage.\\
At the output layer we applied a linear activation function, therefore the network is trained as a regression method rather than classification.
Therefore, we do not quantize $y_{t}$ and use it as continuous
variable. The network is then learned such that the specific
value of $y_{t}$ is predicted from the given input. In this case, a predefined threshold value is used to classify friction values into slippery or non slippery values. To avoid training models that were overfitted, the networks with the lowest evaluation error was stored. Through cross validation we could determine the best parameters for the network.

\section{Results}\label{S:result}

In this section, the performance of the proposed prediction models are evaluated and the effect of different parameters on the performance of the classifiers are investigated.





Table~\ref{t_data} shows the number of samples in three different road segment that are used for the evaluation in this section.
After training the models, we can use them to do prediction for new measurements. All models have been assessed through k-fold cross validation (with k=5), particularly because number of available data samples are limited.
We use error rate, sensitivity, and specificity to access the classification performance, where low values for error rate and high values for sensitivity and specificity are preferred. 
For this purpose, we first compute the true positives (TP, the correctly classified slippery samples), false positives (FP, predicted samples as slippery, when they were non-slippery), true negatives (TN, correctly classified non-slippery samples), and false negatives (FN, predicted samples as non-slippery, when they were slippery). Then, error rate, sensitivity, and specificity are defined as follows: 
\begin{equation}
\begin{split}    
\text{Error rate}&=\frac{FN+FP}{TP+FP+TN+FN},\\
\text{Sensitivity}&=\frac{TP}{TP+FN},\\
\text{Specificity }&=\frac{TN}{TN+FP}.   
\end{split}
\end{equation}

\begin{table}
\centering
\caption{Number of available data samples.}
\resizebox{.8\columnwidth}{!}{
\begin{tabular}{|c|c|c|c|}
\hline 
Road segment  & Number of labeled samples & High friction & Low friction\tabularnewline
\hline 
First & 400 & 119 & 281\\
\hline 
Second & 370 & 271 & 99\\
\hline 
Third & 693 & 370 & 323\\
\hline 
\end{tabular}}
\label{t_data}
\end{table}

\subsection{Effect of the Parameters}
The best number of dimensions to keep after PCA was found to be 14 in our experiments since this gave the best overall results. The following features have been used in the feature vector: combination of previously measured friction values, road surface temperature, dew point temperature, humidity, wiper speed, rain, and snow. 
We did not observe any noticeable improvements of using PCA while training the neural networks, therefore we didn't use it in this case.

As it was discussed earlier, the continuous time data has been discredited into small intervals. To achieve the best results, different time intervals were investigated for down-sampling of data including $\{2,5,10,30,60\}$ minutes. Table~\ref{time_interval} shows the results for the LR method for one road segment, while similar results were obtained for the other methods. As can be seen, using a 2 minute interval results in the least error rate, and therefore in the rest of the paper we only use this interval. 

\begin{table}
\centering
    \caption{Results from varying the time interval for the LR method.}
    \resizebox{.8\columnwidth}{!}{
    \small{
    \begin{tabular}{| c | c | c |c|}
     \hline
     Time intervals& Error rate & Sensitivity & Specificity\\
     \hline
     2 min & 0.2175 & 0.961 & 0.361\\
     5 min & 0.2299& 0.9547 &0.310\\
     10 min& 0.2282  & 0.9569   &0.300\\
     30 min& 0.2530 & 0.9487&  0.245\\
     60 min& 0.2637 &0.9417&  0.244\\
     \hline
    \end{tabular}}}
    \label{time_interval}
\end{table}

To examine the importance of previously measured data points in the feature vector, we evaluate the performance by including data from $\{4,3,2,1\}$ hours ago. Table~\ref{prevhours} shows the results of this experiment for one road segment. In the following, we use the 3 hour alternative as it results to better performance for all methods. 
\begin{table}
\centering
\caption{Results using different duration in the past data as input to the classifiers.}
\resizebox{.8\columnwidth}{!}{
\small{
    \begin{tabular}{|c |c | c | c |c|}
     \cline{2-5} 
     \multicolumn{1}{c|}{}
     &History duration & Error rate & Sensitivity & Specificity\\
     \hline
     \multirow{4}{*}{\rotatebox[origin=c]{90}{LR}} 
     &4 hours& 0.218 & 0.954 &0.378\\
     \cline{2-5}
     &3 hours& 0.223 & 0.954& 0.361\\\cline{2-5}
     &2 hours& 0.243 & 0.940&0.328\\\cline{2-5}
     &1 hours& 0.240 & 0.9431 & 0.328\\
     \hline
     \hline
     \multirow{4}{*}{\rotatebox[origin=c]{90}{SVN}} 
     &4 hours&0.248  & 0.945 & 0.297\\\cline{2-5}
     &3 hours&0.235 & 0.948& 0.333\\\cline{2-5}
     &2 hours&0.227 & 0.941 &0.377\\\cline{2-5}
     &1 hours&0.228 & 0.946 & 0.361\\
     \hline
     \hline
     \multirow{4}{*}{\rotatebox[origin=c]{90}{ANN}} 
     &4 hours& 0.222  & 0.861 &0.581\\\cline{2-5}
     &3 hours&0.220 & 0.875& 0.561 \\\cline{2-5}
     &2 hours&0.237 & 0.865& 0.524\\\cline{2-5}
     &1 hours&0.248 & 0.884& 0.421\\
     \hline
    \end{tabular}}}
    \label{prevhours}
\end{table}

To select the number of hidden layers in the neural network, we evaluate the performance as a function of the hidden layers. Tables~\ref{tab:feature set} shows the results of this experiment. As can be seen, one hidden layer results in the best performance, and it is chosen as the optimal setting for the comparison purposes. One possible reason for this is that we do not have many samples in the data set and it is well known that the performance of neural networks improves by adding more data.

\begin{table}
\centering
    \caption{Results for different numbers of hidden layers used in the Neural network for one road segment.}
    \resizebox{.8\columnwidth}{!}{
    \small{
    \begin{tabular}{| c | c | c |c|}
     \hline
     Hidden layers & Error rate & Sensitivity & Specificity\\
     \hline
     1 & 0.207 & 0.894 & 0.557\\
     2 & 0.215 & 0.855 & 0.629\\
     3 & 0.239 & 0.878 & 0.523\\
     4 & 0.230 & 0.877& 0.527\\
     \hline
    \end{tabular}}}
    \label{tab:feature set}
\end{table}

\subsection{Comparative Results}
Tables~\ref{final_results1},\ref{final_results2}, and~\ref{final_results3} show the result of the classifications methods in three non-overlapping road segments, where each row shows the prediction results in the specified future horizon, given by the minutes.  
Several interesting points can be concluded by studying these results: 

1) The error rate increases slightly for road segment 1 and 2 when the prediction horizon increases. For example, for road segment 2, the error rate increases by around 6\% for the LR method and around 2\% for the ANN method.

2) Considering the interesting case of predicting the friction in the next 30 minutes, ANN and SVM result in comparative  error rates, while their performance regarding sensitivity and specificity are not always consistent. For example, SVM results in higher sensitivity for road segment 1 but much lower specificity for road segment 2.
Overall, ANN leads to more stable results in different conditions.

3) As can be seen,  quite similar error rates are obtained for road segment 1 and 2 while results for road segment 3 are worse. For road segment 1, a high sensitivity is obtained, i.e., most of the slippery samples are correctly classified, while low specificity values are obtained, i.e., many of the non-slippery values are classified as slippery. This behavior is reversed for road segment 2. For road segment 3, the results for different method differ in their sensitivity and specificity and the error rates are higher than the other road segments. 

These results indicate that the final performance depends on the road segment and the used classification method. It should be also noted that in the final application, we might have different preference for sensitivity and specificity. For example, it may be argued that it is more important to have a high sensitivity (i.e., we should not classify any slippery road segment as non-slippery). In this case, ANN will be the winner classification method.

\begin{table}
\caption{Comparative results for road segment 1}
\centering{
\resizebox{.8\columnwidth}{!}{
\small{
\begin{tabular}{|c|c|c|c|c|}
\cline{3-5}
\multicolumn{1}{c}{\multirow{2}{*}{}} &  & Error rate & Sensitivity & Specificity\\
\hline
\multirow{5}{*}{\rotatebox[origin=c]{90}{LR}} 
 & 0 & 0.2250 & 0.9609 & 0.3361 \\
\cline{2-5}
 & 30 & 0.2274 & 0.9312 & 0.4287\\
\cline{2-5}
 & 60 & 0.2298 & 0.9668 & 0.2946 \\
\cline{2-5}
 & 90 & 0.2406 & 0.9430 &  0.3243 \\
\cline{2-5}
 & 120 & 0.2612 & 0.9283 &  0.2857 \\
\hline
\hline
\multirow{5}{*}{\rotatebox[origin=c]{90}{SVM}} 
& 0 & 0.2228 & 0.9358 & 0.4029\\
\cline{2-5}
 & 30 & 0.2226 & 0.9305 & 0.4126\\
\cline{2-5}
 & 60 & 0.2278 & 0.9295& 0.3915\\
\cline{2-5}
 & 90 & 0.2330 & 0.9287 & 0.3838\\
\cline{2-5}
 & 120 & 0.2466 & 0.9122& 0.3738\\
\hline
\hline
\multirow{5}{*}{\rotatebox[origin=c]{90}{ANN}} 
& 0 & 0.1945& 0.9053 & 0.5526\\
\cline{2-5}
 & 30 & 0.2130& 0.8760 & 0.5720\\
\cline{2-5}
 & 60 & 0.2188& 0.8526& 0.5758\\
\cline{2-5}
 & 90 & 0.2299  & 0.8536& 0.5771\\
\cline{2-5}
 & 120 & 0.2303 & 0.8632&0.5666\\
 \hline
\end{tabular}}}
\label{final_results1}
}
\end{table}
\begin{table}
\caption{Comparative results for road segment 2}
\centering{
\resizebox{.8\columnwidth}{!}{
\begin{tabular}{|c|c|c|c|c|}
\cline{3-5}
\multicolumn{1}{c}{\multirow{2}{*}{}} &  & Error rate & Sensitivity & Specificity\\
\hline
\multirow{5}{*}{\rotatebox[origin=c]{90}{LR}} 
 & 0 & 0.2757 & 0.5051 & 0.7823\\
\cline{2-5}
 & 30 & 0.3245  & 0.2556  &0.8474\\
\cline{2-5}
 & 60 & 0.3234 & 0.2000 & 0.8525\\
\cline{2-5}
 & 90 & 0.3271 & 0.1882 &  0.8475\\
\cline{2-5}
 & 120  & 0.3398 & 0.1375 &  0.8428\\
\hline
\hline
\multirow{5}{*}{\rotatebox[origin=c]{90}{SVM}} 
& 0 & 0.2568& 0.1586 & 0.9568\\
\cline{2-5}
 & 30 & 0.2593& 0.0994 &  0.9725\\
\cline{2-5}
 & 60  & 0.2763& 0.0756& 0.9627\\
\cline{2-5}
 & 90  & 0.2551 & 0.1000 & 0.9771\\
\cline{2-5}
 & 120 & 0.2558 & 0.0669& 0.9808\\
\hline
\hline
\multirow{5}{*}{\rotatebox[origin=c]{90}{ANN}} 
& 0 &  0.2648& 0.4384 &0.8460\\
\cline{2-5}
 & 30  & 0.2979& 0.2584 & 0.8621\\
\cline{2-5}
 & 60 & 0.3054& 0.2649 & 0.8521\\
\cline{2-5}
 & 90  & 0.2866  & 0.2276&0.8846 \\
\cline{2-5}
 & 120  & 0.2847 & 0.3525&0.8427\\
 \hline
\end{tabular}}
\label{final_results2}
  }
\end{table}
\begin{table}
\centering{
\caption{Comparative results for road segment 3}
\resizebox{.8\columnwidth}{!}{
\begin{tabular}{|c|c|c|c|c|}
\cline{3-5}
\multicolumn{1}{c}{\multirow{2}{*}{}} &  & Error rate & Sensitivity & Specificity\\
\hline
\multirow{5}{*}{\rotatebox[origin=c]{90}{LR}} 
    &0 & 0.434 & 0.777 & 0.381\\
    \cline{2-5}
     & 30 & 0.467  & 0.766  &0.325\\
     \cline{2-5}
     & 60 & 0.470 & 0.749 & 0.332\\
     \cline{2-5}
     & 90 & 0.476 & 0.793 &  0.266\\
     \cline{2-5}
     & 120 & 0.459 & 0.716 &  0.358\\
\hline
\hline
\multirow{5}{*}{\rotatebox[origin=c]{90}{SVM}} 
  & 0& 0.3576& 0.552 & 0.721\\
  \cline{2-5}
    & 30 & 0.3905& 0.506 & 0.702\\
    \cline{2-5}
    &  60 & 0.3965& 0.540& 0.661\\
    \cline{2-5}
     & 90 & 0.3949 & 0.523 & 0.684\\
     \cline{2-5}
    &  120 & 0.3686 & 0.654& 0.608\\
\hline
\hline
\multirow{5}{*}{\rotatebox[origin=c]{90}{ANN}} 
    & 0 & 0.365 & 0.557 & 0.713\\
     \cline{2-5}
    &  30 & 0.349& 0.570 &0.724\\
     \cline{2-5}
     & 60 & 0.357& 0.554&0.724\\
      \cline{2-5}
     & 90 & 0.374  & 0.570&0.679\\
      \cline{2-5}
    &  120 & 0.355 & 0.653&0.643\\
 \hline
\end{tabular}}
\label{final_results3}
}
\end{table}
\section{Conclusion}\label{S:conclusion}
Three classification models are proposed to predict the friction level in road segments using history data. 
The proposed classification methods (logistic regression, support vector machine, and artificial neural network) are evaluated under different setting including forecast time horizon, feature vector, and number of hidden layers, and for each method error rate, sensitivity, and specificity are reported.
Data is collected from fleet of cars and weather stations at the city of Gothenburg in Sweden. Our experiments show that an error rate in the order of 20-30 $\%$ can be obtained while the prediction accuracy slightly changes for different road segments. Although no single method leads to the best results in all conditions, the ANN method results in more stable results considering different conditions. 

\bibliographystyle{IEEEtran}
\tiny{
\bibliography{NasserRefs}}

\end{document}